\documentclass{llncs}
\usepackage{latexsym,amsmath,amssymb,amsfonts}
\usepackage{url,color}
\usepackage{tikz} 

\newcommand\pr[2]{\mathop{#1\vert_{#2}}}

\begin{document}

\title{On First-Order Model-Based Reasoning}

\author{Maria Paola Bonacina\inst{1}
\and%
Ulrich Furbach\inst{2}
\and%
Viorica Sofronie-Stokkermans\inst{2}}

\institute{Dipartimento di Informatica,
Universit\`a degli Studi di Verona, Italy\\
\email{mariapaola.bonacina@univr.it}
\vskip 6pt
\and
Fachbereich Informatik,
Universit\"at Koblenz-Landau, Germany\\
\email{furbach@uni-koblenz.de},
\email{sofronie@uni-koblenz.de}
}
%
\maketitle
\begin{quote}
\begin{flushright}
Dedicated to\\
{\bf Jos\'e Meseguer},\\
friend and colleague.
\end{flushright}
\end{quote}

\begin{abstract}
Reasoning semantically in first-order logic is notoriously a challenge.
This paper surveys a selection of {\em semantically-guided} or {\em model-based}
methods that aim at meeting aspects of this challenge.
For first-order logic we touch upon {\em resolution-based} methods,
{\em tableaux-based} me\-thods, {\em DPLL-inspired} methods,
and we give a preview of a new method called SGGS,
for {\em Semantically-Guided Goal-Sensitive} reasoning.
For first-order theories we highlight {\em hierarchical} and {\em locality-based} methods,
concluding with the recent {\em Model-Constructing satisfiability calculus}.
\end{abstract}

\section{Introduction}

Traditionally,
automated reasoning has centered on {\em proofs} rather than {\em models}.
However, models are useful for applications, intuitive for users,
and the notion that {\em semantic guidance}
would help proof search is almost as old as
theorem proving itself.
In recent years there has been a surge of {\em model-based}
first-order reasoning methods,
inspired in part by the success of model-based
solvers for propositional satisfiability (SAT)
and satisfiability modulo theories (SMT).

The core procedure of these solvers is
the {\em conflict-driven clause learning} (CDCL) version
\cite{MSA:ICCAD1996,MMZZM:DAC2001,LZ-SM:CADE2002,MZ:CACM:2009}
of the Davis-Putnam-Logemann-Loveland (DPLL) procedure for propositional logic
\cite{DPLL:CACM:1962}.
The original Davis-Putnam (DP) procedure \cite{DP:JACM:1960}
was proposed for first-order logic,
and featured propositional, or ground, resolution.
The DPLL procedure replaced propositional resolution with {\em splitting},
initially viewed as breaking disjunctions apart by {\em case analysis},
to avoid the growth of clauses and the non-determinism of resolution.
Later, splitting was understood as {\em guessing}, or {\em deciding},
the truth value of a propositional variable,
in order {\em to search for a model} of the given set of clauses.
This led to read DPLL as a {\em model-based} procedure,
where all operations are centered around a candidate partial model,
called {\em context},
represented by a sequence, or {\em trail}, of literals.

DPLL-CDCL brought back propositional resolution as a mechanism
to generate {\em lemmas},
and achieve a better balance between {\em guessing} and {\em reasoning}.
The model-based character of the procedure became even more pronounced:
when the current candidate model falsifies a clause,
this {\em conflict} is {\em explained} by a heuristically controlled series of resolution steps,
a resolvent is added as lemma,
and the candidate partial model is repaired in such a way to remove the conflict,
satisfy the lemma,
and backjump as far away as possible from the conflict.
SMT-solvers integrate in DPLL-CDCL a decision procedure for satisfiability in
a theory or combination of theories ${\cal T}$:
the ${\cal T}$-satisfiability procedure raises a {\em ${\cal T}$-conflict}
when the candidate partial model is not consistent with ${\cal T}$,
and generates {\em ${\cal T}$-lemmas} to add theory reasoning
to the inference component \cite{SMTsurvey:2009,dMB:CACM:2011}.

While SAT and SMT-solvers offer fast decision procedures,
they typically apply to sets of propositional or ground clauses,
without quantifiers.
Indeed, decidability of the problem and termination of the procedure descend from the fact
that the underlying language is the {\em finite} set of the input atoms.

ATP (Automated Theorem Proving) systems offer theorem-proving strategies
that are designed for the far more expressive language of first-order logic,
but are only {\em semi-decision procedures} for validity,
as the underlying language, and search space, are {\em infinite}.
This trade-off between {\em expressivity} and {\em decidability} is ubiquitous
in logic and artificial intelligence.
First-order satisfiability is not even semi-decidable,
which means that first-order model-building cannot be mechanized in general.
Nevertheless,
there exist first-order reasoning methods that are
{\em semantically-guided} by a {\em fixed} interpretation,
and even {\em model-based},
in the sense that the state of a derivation
contains a representation of a candidate partial model
that evolves with the derivation.

In this survey,
we illustrate a necessarily incomplete selection of such methods for first-order logic
(Section~\ref{mbr-fol}) or first-order theories (Section~\ref{mbr-th}).
In each section the treatment approximately goes
from syntactic or axiomatic approaches towards more semantic ones,
also showing connections with Jos\'e Meseguer's work.
All methods are described in expository style, and
the interested reader may find the technical details in the references.
Background material is available in previous surveys,
such as
\cite{Plaisted:1990,Plaisted:1993,MPB:LNAI1600:1999:taxonomy,LMP:HB:KRR:2008,Plaisted:2014}
for theorem-proving strategies,
\cite{MPB:PPDP2010:OnTPforPC} for decision procedures
based on theorem-proving
strategies or their integration with SMT-solvers,
and books such as
\cite{DP-YZ:book:1997,Bibel:Schmitt:1998,DBLP:books/el/RobinsonV01}.

\section{Model-based Reasoning in First-Order Logic}\label{mbr-fol}

In this section we cover
{\em semantic resolution},
which represents the early attempts at injecting semantics in resolution;
{\em hypertableaux},
which illustrates model-based reasoning in tableaux,
with applications to fault diagnosis and description logics;
the {\em model-evolution calculus}, which lifts DPLL to first-order logic,
and a new method called {\em SGGS},
for {\em Semantically-Guided Goal-Sensitive} reasoning,
which realizes a first-order CDCL mechanism.

\subsection{Semantic Resolution}

Soon after the seminal article by Alan Robinson
introducing the resolution principle \cite{Robinson:1965:a},
James\,R.\,Slagle presented \textit{semantic resolution} in \cite{Slagle:1967}.
Let $S$ be the finite set of first-order clauses to be refuted.
Slagle's core idea was to use a given Herbrand interpretation $I$
to avoid generating resolvents that are true in $I$,
since expanding a consistent set should not lead to a refutation.
The following example from \cite{ChangLee} illustrates the concept
in propositional logic:

\begin{example}\label{ex:semanticResolution:prop}
Given $S = \{\neg A_1\vee \neg A_2\vee A_3,
\ A_1\vee A_3,
\ A_2\vee A_3\}$,
let $I$ be all-negative,
that is, $I = \{\neg A_1, \neg A_2, \neg A_3\}$.
Resolution between $\neg A_1\vee \neg A_2\vee A_3$ and
$A_1\vee A_3$ generates $\neg A_2\vee A_3$,
after merging identical literals.
Similarly,
resolution between $\neg A_1\vee \neg A_2\vee A_3$ and
$A_2\vee A_3$ generates $\neg A_1\vee A_3$.
However, these two resolvents are true in $I$.
Semantic resolution prevents generating such resolvents,
and uses all three clauses to generate only $A_3$,
which is false in $I$.
\end{example}

Formally, say that we have a clause $N$, called \textit{nucleus},
and clauses $E_1, \ldots, E_q$, with $q\geq 1$,
called \textit{electrons},
such that the electrons are {\em false} in $I$.
Then, if there is a series of clauses
$R_1,\ R_2,\ldots,R_q,\ R_{q+1}$,
where $R_1$ is $N$,
$R_{i+1}$ is a resolvent of $R_i$ and $E_i$,
for $i = 1, \ldots, q$,
and $R_{q+1}$ is {\em false} in $I$,
semantic resolution generates only $R_{q+1}$.
The intuition is that electrons are used to resolve away literals
in the nucleus until a clause false in $I$ is generated.

\begin{example}
In the above example,
$\neg A_1\vee \neg A_2\vee A_3$ is the nucleus $N$,
and $A_1\vee A_3$ and
$A_2\vee A_3$ are the electrons $E_1$ and $E_2$, respectively.
Resolving $N$ and $E_1$ gives $\neg A_2\vee A_3$,
and resolving the latter with $E_2$ yields $A_3$:
only $A_3$ is retained, while
the intermediate resolvent $\neg A_2\vee A_3$ is not.
\end{example}

Semantic resolution can be further restricted by assuming
a precedence $>$ on predicate symbols,
and stipulating that
in each electron the predicate symbol of the literal resolved upon
must be maximal in the precedence.
The following example also from \cite{ChangLee} is in 
first-order logic:

\begin{example}\label{ex:semanticResolution:fol}
For $S = \{Q(x)\vee Q(a)\vee \neg R(y)\vee \neg R(b)\vee S(c),
\ \neg Q(z)\vee \neg Q(a),
\ R(b)\vee S(c)\}$,
let $I$ be $\{Q(a), Q(b), Q(c), \neg R(a), \neg R(b), \neg R(c),
\neg S(a), \neg S(b),\neg S(c)\}$,
so that $I\not\models \neg Q(z)\vee \neg Q(a)$ and
$I\not\models R(b)\vee S(c)$.
Assume the precedence $Q > R > S$.
Thus, $Q(x)\vee Q(a)\vee \neg R(y)\vee \neg R(b)\vee S(c)$
is the nucleus $N$,
and $\neg Q(z)\vee \neg Q(a)$ and $R(b)\vee S(c)$
are the electrons $E_1$ and $E_2$, respectively.
Resolution between $N$ and $E_1$ on the $Q$-literals
produces $\neg R(y)\vee \neg R(b)\vee S(c)$,
which is not false in $I$, and therefore it is not kept.
Note that this resolution step is a binary resolution step between
a factor of $N$ and a factor of $E_1$.
Resolution between $\neg R(y)\vee \neg R(b)\vee S(c)$
and $E_2$ on the $R$-literals yields $S(c)$.
This second resolution step is a binary resolution between a factor
of $\neg R(y)\vee \neg R(b)\vee S(c)$ and $E_2$.
Resolvent $S(c)$ is false in $I$ and it is kept.
\end{example}

In these examples $I$ is given by a finite set of literals:
Example~\ref{ex:semanticResolution:prop} is propositional,
and in Example~\ref{ex:semanticResolution:fol} the Herbrand base is finite,
because there are no function symbols.
The examples in \cite{Slagle:1967} include a theorem from algebra,
where the interpretation is given by a multiplication table
and hence is really of semantic nature.
The crux of semantic resolution is the {\em representation} of $I$.
In theory,
a Herbrand interpretation is given
by a subset of the Herbrand base of $S$.
In practice,
one needs a {\em finite} representation of $I$,
which is a non-trivial issue,
whenever the Herbrand base is not finite,
or a mechanism to test the truth of a literal in $I$.
Two instances of semantic resolution that aimed at addressing
this issue are \textit{hyperresolution} \cite{Robinson:1965:b}
and the \textit{set-of-support strategy} \cite{wos}.

Hyperresolution assumes that $I$ contains
either all negative literals
or all positive literals.
In the first case,
it is called \textit{positive} hyperresolution,
because electrons and all resolvents are positive clauses:
positive electrons are used to resolve away all negative literals
in the nucleus to get a positive hyperresolvent.
In the second case,
it is called \textit{negative} hyperresolution,
because electrons and all resolvents are negative clauses:
negative electrons are used to resolve away all positive literals
in the nucleus to get a negative hyperresolvent.
Example~\ref{ex:semanticResolution:prop} is an instance
of positive hyperresolution.

The set-of-support strategy assumes that $S = T \uplus SOS$,
where $SOS$ (for Set of Support)
contains initially the clauses coming from the negation of the conjecture,
and $T = S \setminus SOS$ is consistent,
for some $I$ such that $I\models T$ and $I\not\models SOS$.
A resolution of two clauses is permitted,
if at least one is from $SOS$,
in order to avoid expanding the consistent set $T$.
All resolvents are added to $SOS$.
Thus, all inferences involve clauses
descending from the negation of the conjecture:
a method with this property is deemed {\em goal-sensitive}.

In terms of implementation,
positive hyperresolution is often implemented in
contemporary theorem pro\-vers by resolution with
{\em selection of negative literals}.
Indeed,
resolution can be restricted by a {\em selection function}
that selects negative literals \cite{BacGan:JLC:1994}.
A clause can have all, some, or none of its negative literals selected,
depending on the selection function.
In {\em resolution with negative selection},
the negative literal resolved upon must be selected,
and the other parent must not contain selected literals.
If some negative literal is selected for each clause containing one,
one parent in each resolution inference will be a positive clause,
that is, an electron for positive hyperresolution.
Thus,
a selection function that selects some negative literal
in each clause containing one
induces resolution to simulate hyperresolution
as a macro inference involving several steps of resolution.

The set-of-support strategy is available in all theorem provers
that feature the {\em given-clause loop} \cite{otter3},
which is a de facto standard for resolution-based provers.
This algorithm maintains two lists of clauses,
named {\em to-be-selected} and {\em already-selected},
and at each iteration it extracts a {\em given clause}
from {\em to-be-selected}.
In its simplest version,
with only resolution as inference rule,
it performs all resolutions between the given clause and the clauses
in {\em already-selected};
adds all resolvents to {\em to-be-selected};
and adds the given clause to {\em already-selected}.
If one initializes these lists by putting
the clauses in $T$ in {\em already-selected},
and the clauses in $SOS$ in {\em to-be-selected},
this algorithm implements the set-of-support strategy.
Indeed, in the original version of the given-clause algorithm,
{\em to-be-selected} was called {\em SOS},
and {\em already-selected} was called {\em Usable}.

State-of-the-art resolution-based theorem provers
implement more sophisticated versions of the given clause algorithm,
which also accomodate {\em contraction rules},
that delete (e.g., {\em subsumption}, {\em tautology deletion})
or simplify clauses (e.g., {\em clausal simplification},
{\em equational simplification}).
The compatibility of contraction rules with semantic strategies
is not obvious, as shown by the following:

\begin{example}
Let $T = \{\neg P,\ P\vee Q\}$ and $SOS = \{\neg Q\}$.
Clausal simplification,
which is a combination of resolution and subsumption,
applies $\neg Q$ to simplify $P\vee Q$ to $P$.
If the result is $T = \{\neg P,\ P\}$ and $SOS = \{\neg Q\}$,
the consistent set $T$ becomes inconsistent,
and the refutational completeness of resolution with set-of-support
collapses, since the set-of-support strategy does not allow us to resolve
$P$ and $\neg P$, being both in $T$.
The correct application of clausal simplification yields
$T = \{\neg P\}$ and $SOS = \{\neg Q,\ P\}$,
so that the refutation can be found.
\end{example}

In other words,
if a clause in $SOS$ simplifies a clause,
whether in $T$ or in $SOS$,
the resulting clause must be added to $SOS$.
The integration of contraction rules
and other enhancements, such as {\em lemmaizing},
in semantic strategies
was investigated in general in \cite{MPB-JH:NGC:1998:semrlc}.

Semantic resolution,
hyperresolution, and the set-of-support strategy
exhibit {\em semantic guidance}.
We deem a method {\em semantically guided},
if it employs a {\em fixed} interpretation to drive the inferences.
We deem a method {\em model-based},
if it builds and transforms a {\em candidate model},
and uses it to drive the inferences.

A beginning of the evolution from being semantically guided
to being model-based
can be traced back to the SCOTT system \cite{slaney1994scott},
which combined the finite model finder FINDER,
that searches for small models,
and the resolution-based theorem prover OTTER \cite{otter3}.
As the authors write
``SCOTT brings semantic information gleaned from the proof attempt
into the service of the syntax-based theorem prover.''
In SCOTT, FINDER provides OTTER with a {\em guide model},
which is used for an extended set-of-support strategy:
in each resolution step at least one of the parent clauses
must be false in the guide model.
During the proof search FINDER updates periodically
its model to make more clauses true.
Thus, inferences are controlled as in the set-of-support strategy,
but the guide model is {\em not fixed}, which is why
SCOTT can be seen as a forerunner of model-based methods.
Research on the cooperation between theorem prover and finite model finder
continued with successors of OTTER, such as Prover9,
and successors of FINDER, such as MACE4 \cite{ZZinMcCuneBook}.
This line of research has been especially fruitful in
applications to mathematics
(e.g., \cite{ArthanOlivainMcCuneBook,FitelsoninMcCuneBook}).

\subsection{Hypertableaux}\label{subsect:hypertab}

Tableau calculi offer an alternative to resolution
and they have been discussed abundantly in the literature
(e.g., Chapter 3 in \cite{DBLP:books/el/RobinsonV01}).
Their advantages include no need for a clause normal form,
a single proof object, and an easy extendability to other logics.
The disadvantage, even in the case of clause normal form tableaux,
is that variables are {\em rigid},
which means that substitutions have to be applied to all occurrences of a
variable within the entire tableau.
The {\em hypertableau calculus} \cite{baumgartner1996hyper}
offers a more liberal treatment of variables,
and borrows the concept of hyperinference from positive hyperresolution. 

In this section,
we adopt a Prolog-like notation for clauses:
$A_1 \lor \ldots\lor A_m \lor \lnot  B_1 \lor \ldots\lor \lnot B_n$
is written $A_1, \ldots, A_m \Leftarrow B_1, \ldots, B_n$,
where $A_1, \ldots, A_m$ form the {\em head} of the clause
and are called {\em head literals},
and $B_1, \ldots, B_n$ form the {\em body}.
There are two rules for constructing a hypertableau
(cf. \cite{baumgartner1996hyper}):
the \textit{initialization} rule
gives a tableau consisting of a single node labeled with $\top$;
this one-element branch is \textit{open}.
The \textit{hyperextension} rule selects an open branch and a clause
$A_1, \ldots, A_m \Leftarrow B_1, \ldots, B_n$,
where $m,n \geq 0$,
from the given set $S$,
such that there exists a most general unifier $\sigma$
which makes all the $B_i \sigma$'s {\em follow logically} from the model
given by the branch.
If there is a variable in the clause
that has an occurrence in more than one head literal $A_i$,
a \textit{purifying substitution} $\pi$ is used to ground this variable.
Then the branch is extended by new nodes labeled with
$A_i\sigma\pi, \ldots, A_m\sigma\pi$.
A branch is {\em closed} if it can be extended by a clause
without head literals.
$S$ is unsatisfiable if and only if
there is a hypertableau for $S$ whose branches are all closed.

Two major advantages of hyperextension are that
it avoids unnecessary branching,
and only variables in the clauses are universally quantified
and get instantiated,
while variables in the branches are treated as {\em free} variables
(except those occurring in different head literals).
The latter feature allows a su\-per\-po\-si\-tion-\-like handling of equality
\cite{Baumgartner-Furbach-Pelzer:JLC:2008:HyperTableauxEq},
while the former is relevant for hypertableaux for description logic
\cite{smh08HermiT},
which we shall return to in the next section.
Hypertableaux were implemented in the {\em Hyper}
theorem prover for first-order logic,
followed by {\em E-Hyper} implementing also the handling of equality.

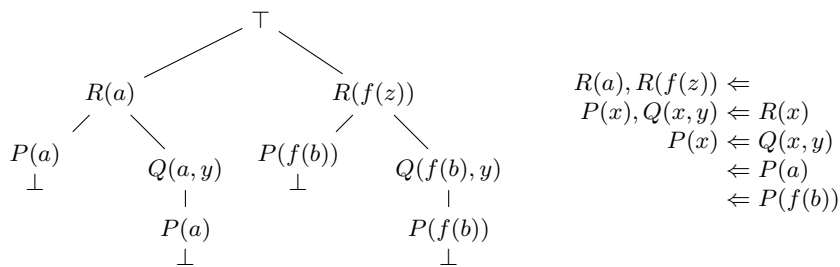
\begin{figure}[hbtp]
\begin{minipage}{5cm}
\begin{tikzpicture}
\node[name = n1,anchor=south]{
$\top$
   
}
        [sibling distance = 4.0cm, level distance=1.0cm, node distance = 3cm] 
        child[sibling distance =4.0cm, node distance = 3cm]{
            node {$R(a)$}  
 [sibling distance = 2.0cm, node distance = 3cm, level distance=1cm]
            child{
                node {$\begin{array}{c} P(a)\\ \bot \end{array}$}
                        [sibling distance = 2.cm, node distance = 3cm, level distance=1cm]
                }
            child{
                node{$Q(a,y)$}
                [sibling distance = 2.cm, node distance = 3cm,level distance=1cm]    	
            		child{
                		node{$\begin{array}{c} P(a) \\ \bot \end{array}$}
                	}
                }          
        }                      
        child[sibling distance = 3.cm, level distance=1.0cm, node distance = 3cm]{
            node {$R(f(z))$}
        [sibling distance = 2.0cm, node distance = 3cm, level distance=1cm]
            child{
                node {$\begin{array}{c} P(f(b)) \\ \bot \end{array}$}
                        [sibling distance = 2.cm, node distance = 3cm, level distance=1cm]
                }
            child{
                node{$Q(f(b),y)$}
                [sibling distance = 2.cm, node distance = 3cm,level distance=1cm]    	
            		child{
                		node{$\begin{array}{c} P(f(b)) \\ \bot \end{array}$}
                	}
                }
        };
\end{tikzpicture}
\end{minipage}
\hskip 2.5 cm
\begin{minipage}{3 cm}
\begin{array}[b]{rll}
R(a), R(f(z)) & \Leftarrow &\\
P(x), Q(x,y) & \Leftarrow & R(x)\\
P(x) & \Leftarrow & Q(x,y)\\
 & \Leftarrow & P(a)\\
 & \Leftarrow & P(f(b))
\end{array}
\end{minipage}
\caption[Example Refutation]{\small A sample hypertableaux refutation with
the clause set on the right.}\label{F:refexample}
\end{figure}

\begin{example}
An example refutation is given in Figure~\ref{F:refexample}.
The initial tableau is set up with the only positive clause.
Extension at $R(a)$ with the second clause uses $\sigma = \{x\gets a\}$:
since $y$ appears only once in the resulting head,
$\pi  = \varepsilon$ and $y$ remains as a free variable.
In the right subtree $R(f(z))$ is extended with the second clause
and $\sigma = \{x\gets f(z)\}$.
In the head $P(f(z)), Q(f(z),y)$ of the resulting clause $z$ is repeated:
an {\em instance generation} mechanism produces $\pi = \{ z \gets b\}$,
or the instance $P(f(b)), Q(f(b),y) \Leftarrow R(f(b))$,
to find a refutation.
Note how the tableau contains by construction only positive literals, and
the interpretation given by a branch is used to control the extension steps
very much like in hyperresolution. 
\end{example}

\subsection{Model-based Transformation of Clause Sets}

Hypertableaux use partial models,
that is, models for parts of a clause set,
to control the search space.
An open branch that cannot be expanded further
represents a model for the entire clause set.
In this section we present a transformation method,
borrowed from model-based diagnosis and presented in
\cite{baumgartner1997semantically},
which is based on a given model
and therefore can be installed on top of hypertableaux.
In applications to diagnosis,
one has a set of clauses $S$ which corresponds to a description of a system,
such as an electrical circuit.
Very often there is a model $I$ of a correctly functioning system available;
in case of an electrical circuit it may be provided by the design tool itself.
If the actual circuit is fed with an input
and does not show the expected output,
the task is to find a diagnosis,
or those parts of the circuit which may be broken.
Instead of doing reasoning with the system description $S$
and its input and output in order to find the erroneous parts,
the idea is to compute only {\em deviations}
from the initially given model $I$. 

Assume that $S$ is a set of propositional clauses
and $I$ a set of propositional atoms; as a very simple example take 
\[S = \{B \Leftarrow, \; C \Leftarrow A, B\} \textrm{ and } I = \{A\}.\]
Each clause in $S$ is transformed by replacing a positive literal $L$
by $\neg neg\_L$ and a negative literal $\neg L$ by $neg\_L$,
if $L$ is contained in $I$.
In other words,
a literal which is contained in the initial model moves
to the other side of the arrow
and is renamed with the prefix $neg\_$ as in 
\[S^\prime = \{B \Leftarrow, \; C, neg\_A \Leftarrow B\}.\]

This transformation is model-preserving,
as every model of $S$ is a model of $S^\prime$.
For this it suffices to assign true to $neg\_L$
if and only if $L$ is false,
for every $L \in I$,
and keep truth values unchanged for atoms outside of $I$.
This property is independent of $I$,
and it holds even if $I$ is not a model of $S$.
In our example,
after initialization,
first hyperextension with $B \Leftarrow$,
and then hyperextension with $C, neg\_A \Leftarrow B$,
yield the open branches $\{B, C\}$ and $\{B,neg\_A \}$.
Hyperextension with $C, neg\_A \Leftarrow B$ can be applied
because only $B$ occurs in the body.
Since $A$ is assumed to be true in $I$,
it can be added:
adding $A$ to $\{B, C\}$ yields model $\{A,B,C\}$;
adding $A$ to $\{B,neg\_A \}$ yields model $\{B\}$.
If deriving $A$ in $S$ is very expensive,
it pays off to save this derivation by moving $A$ as $neg\_A$
to the body of the clause.
In this example a Horn clause becomes non-Horn,
introducing the case where $A$ is false,
and $neg\_A$ holds,
although $A$ is in $I$.
Symmetrically, a non-Horn clause may become Horn.
This transformation technique enabled a hypertableau prover
to compute benchmarks from electrical engineering
\cite{baumgartner1997semantically},
and was also applied to the view update problem
in databases \cite{AravindanBaumgartner:JSC:2000}.

Although this transformation mechanism only works in the propositional case,
it can be extended to description logic \cite{FurbachSchon:TABLEAUX2013}.
Indeed,
most description logic reasoners are based on tableau calculi,
and a hypertableau calculus was used in \cite{smh08HermiT} as a basis
for an efficient reasoner for the description logic $\mathcal{SHIQ}$.
For this purpose,
the authors define \textit{DL-clauses}
as clauses without occurrences of function symbols,
and such that the head is allowed to include disjunctions of atoms,
which may contain existential role restrictions as in 
\[\exists \mathit{repairs.Car}(x)\Leftarrow \mathit{Mechanic(x).} \]
In other words,
a given $\mathcal{SHIQ}$-Tbox is translated to a large extent
into first-order logic;
only existential role restrictions are kept as positive ``literals.''
Given a Tbox in the form of a set of DL-clauses,
if we have in addition an Abox, or a set of ground assertions,
we can use the interpretation given by the ABox
as initial model for the model-based transformation \cite{FurbachSchon:TABLEAUX2013}.
On this basis,
the already mentioned {\em E-Hyper} reasoner was modified
to become {\em E-KRHyper},
which was shown to be a decision procedure for $\mathcal{SHIQ}$
in \cite{BPS:CADE2013}. 
\subsection{The Model Evolution Calculus}

The practical success of DPLL-based SAT solvers suggested the goal
of lifting features of DPLL to the first-order level.
Research focused on {\em splitting first-order clauses},
seen as a way to improve the capability to handle non-Horn clauses.
Breaking first-order clauses apart is not as simple as in propositional logic,
because a clause stands for all its ground instances,
and literals share variables that are implicitly universally quantified.
Decomposing disjunction is a native feature in tableaux,
whose downside is represented by rigid variables,
as already discussed in Section~\ref{subsect:hypertab},
where we saw how hypertableaux offer a possible answer.

The quest for ways to split efficiently clauses
such as $A(x) \vee B(x)$ led to
the {\em model evolution calculus} \cite{MEC:AI:2008}.
In this method
splitting $A(x) \vee B(x)$ yields a branch with $A(x)$,
meaning $\forall x A(x)$,
and one with $\neg A(c)$,
the Skolemized form of $\neg \forall x A(x) \equiv \exists x \neg A(x)$.
Splitting in this way has the disadvantage that the signature changes,
and Skolem constants, being new, do not unify with other non-variable terms.
Thus, the model evolution calculus employs {\em parameters},
in place of Skolem constants,
to replace existentially quantified variables.
These parameters are similar to the free variables of hypertableaux.

The similarity between the model evolution calculus and DPLL
goes beyond splitting,
as the model evolution calculus aims at being a faithful lifting
of DPLL to first-order logic.
Indeed, a central feature of
the model evolution calculus is that it maintains a {\em context} $\Lambda$,
which is a finite set of literals,
representing a Herbrand interpretation $I_{\Lambda}$,
seen as a candidate partial model of the input set of clauses $S$.
Thus, the model evolution calculus
is a {\em model-based} first-order method.
Literals in $\Lambda$ may contain variables,
implicitly universally quantified as in clauses,
and parameters.
Clauses are written in the form $\Lambda \vdash C$,
so that each clause carries the context with itself.

In order to determine whether 
$I_{\Lambda}\models L$, for $L$ an atom in the Herbrand base of $S$,
one looks at the most specific literal in $\Lambda$ that subsumes $L$;
in case of a tie, $L$ is picked with positive sign.
If $I_{\Lambda}$ is not a model of $S$,
the inference system unifies input clauses against $\Lambda$
to find instances that are not true in $I_{\Lambda}$:
these instances are subject to splitting,
to modify $\Lambda$ and repair $I_{\Lambda}$.
Otherwise, the system recognizes that $\Lambda$ cannot be fixed and
declares $S$ unsatisfiable.
As DPLL uses {\em depth-first search with backtracking},
the model evolution calculus uses depth-first search with backtracking
and {\em iterative deepening} on term depth,
which however may skew the search
towards big proofs with small term depth.
The model evolution calculus was implemented in the {\em Darwin} prover
\cite{MEC:IJAI:2006},
and extended to handle equality on its own \cite{MEE:JSC:2012}
and with superposition \cite{MEC+SUP:CADE2009}.

\subsection{SGGS: Semantically-Guided Goal-Sensitive Reasoning}\label{subsect:sggs}

SGGS, for {\em Semantically-Guided Goal-Sensitive} reasoning,
is a new theorem-proving method for first-order logic
\cite{MPB-DAP:PAAR2014:SGGSexpository,MPB-DAP:UNIF2014:SGGSconstraints,MPB-DAP:JAR1:SGGSmodel,MPB-DAP:JAR2:SGGSinference},
which inherits features from several of the strategies
that we surveyed in the previous sections.
SGGS is {\em semantically guided} by a fixed initial interpretation $I$
like semantic resolution;
and it is {\em goal-sensitive} like the set-of-support strategy.
With hyperresolution and hypertableaux,
it shares the concept of hyperinference,
although the hyperinference in SGGS, as we shall see,
is an instance generation inference,
and therefore its closest ancestor is
{\em hyperlinking} \cite{SJL-DP:JAR:1992,DP-YZ:JAR:2000},
an inference rule that uses the most general unifier of a
hyperresolution step to generate instances of the parents,
rather than a hyperresolvent.

Most importantly,
SGGS is {\em model-based} at the first-order level,
in the sense of working by representing and transforming
a candidate partial model of the given set $S$ of first-order clauses.
This fundamental characteristic is in common with
the model evolution calculus,
but while the latter lifts DPLL,
SGGS lifts DPLL-CDCL to first-order logic,
and it combines the model-based character
with the semantic guidance and the goal sensitivity.
Indeed, SGGS was motivated by the quest for a method
that is simultaneously first-order, model-based,
semantically-guided, and goal-sensitive.
Furthermore,
SGGS is {\em proof confluent},
which means it does not need backtracking,
and it does not necessarily reduce to either DPLL or DPLL-CDCL,
if given a propositional problem.

In DPLL-CDCL,
if a literal $L$ appears in the trail
that represents the candidate partial model,
all occurrences of $\neg L$ in the set of clauses are false.
If all literals of a clause $C$ are false,
$C$ is in {\em conflict};
if all literals of $C$ except one, say $Q$, are false,
$Q$ is an {\em implied literal} with $C$ as {\em justification}.
The status of $C$ {\em depends} on the {\em decision levels}
where the complements of its literals were either guessed
(decision) or implied (Boolean propagation).
SGGS generalizes these concepts to first-order logic.
Since variables in first-order literals are implicitly universally quantified,
if $L$ is true, $\neg L$ is false,
but if $L$ is false, we only know that a ground instance of $\neg L$ is true.
SGGS restores the symmetry by introducing the notion of {\em uniform falsity}:
$L$ is uniformly false, if all its ground instances are false,
or, equivalently, if $\neg L$ is true.
A first r\^ole of the given interpretation $I$ is to provide a
{\em reference model} where to evaluate the truth value of literals:
a literal is {\em $I$-true}, if it is true in $I$,
and {\em $I$-false}, if it is uniformly false in $I$.

An {\em SGGS clause sequence} $\Gamma$ is a sequence of clauses,
where every literal is either $I$-true or $I$-false,
so that it tells the truth value in $I$ of all its ground instances.
In every clause $C$ in $\Gamma$ a literal is {\em selected}:
if $C = L_1\vee\ldots\vee L_n$ and $L_n$ is selected,
we write the clause as $L_1\vee\ldots\vee [L_n]$,
or, more compactly, $C[L_n]$, with a slight abuse of the notation.
SGGS tries to modify $I$ into a model of $S$
(if $I$ is a model of $S$ the problem is solved).
Thus, $I$-false literals are preferred for selection,
and an $I$-true literal is selected only in
a clause whose literals are all $I$-true, called {\em $I$-all-true} clause.
A second r\^ole of the given interpretation $I$ is to provide a
{\em starting point} for the search of a model for $S$.

An SGGS clause sequence $\Gamma$ represents
a {\em partial interpretation} $I^p(\Gamma)$:
if $\Gamma$ is the empty sequence,
denoted by $\varepsilon$,
$I^p(\Gamma)$ is empty;
if $\Gamma$ is $C_1[L_1],\ldots,C_i[L_i]$,
and $I^p(\pr{\Gamma}{i-1})$ is the partial interpretation represented
by $C_1[L_1],\ldots,C_{i-1}[L_{i-1}]$,
then $I^p(\Gamma)$ is $I^p(\pr{\Gamma}{i-1})$
plus the ground instances $L_i\sigma$ of $L_i$,
such that $C_i\sigma$ is ground,
$C_i\sigma$ is not satisfied by $I^p(\pr{\Gamma}{i-1})$,
and $\neg L_i\sigma$ is not in $I^p(\pr{\Gamma}{i-1})$,
so that $L_i\sigma$ can be added to satisfy $C_i\sigma$.
In other words,
each clause adds the ground instances of its selected literal
that satisfy ground instances of the clause not satisfied thus far.

An {\em interpretation} $I[\Gamma]$ is obtained
by consulting first $I^p(\Gamma)$,
and then $I$:
for a ground literal $L$,
if its atom appears in $I^p(\Gamma)$,
its truth value in $I[\Gamma]$ is that in $I^p(\Gamma)$;
otherwise, it is that in $I$.
Thus, $I[\Gamma]$ is $I$ modified to satisfy the clauses in $\Gamma$
by satisfying the selected literals,
and since $I$-true selected literals are already true in $I$,
the $I$-false selected literals are those that matter.
For example,
if $\Gamma$ is
$[P(x)],\ \neg P(f(y))\vee [Q(y)],\ \neg P(f(z))\vee \neg Q(g(z))\vee [R(f(z),g(z))]$,
and $I$ is all negative like in positive hyperresolution,
$I[\Gamma]$ satisfies all ground instances of
$P(x)$, $Q(y)$, and $R(f(z),g(z))$,
and no other positive literal.

SGGS generalizes Boolean, or clausal, propagation to first-order logic.
Consider an $I$-false ($I$-true) literal $M$ selected in clause $C_j$
in $\Gamma$,
and an $I$-true ($I$-false) literal $L$ in $C_i$, $i > j$:
if all ground instances of $L$ appear negated
among the ground instances of $M$ added to $I^p(\Gamma)$,
$L$ is uniformly false in $I[\Gamma]$ because of $M$,
and {\em depends} on $M$,
like $\neg L$ {\em depends} on $L$
in propositional Boolean propagation,
when $L$ is in the trail.
If this happens for {\em all} its literals,
clause $C[L]$ is in {\em conflict} with $I[\Gamma]$;
if this happens for all its literals except $L$, 
$L$ is an {\em implied literal} with $C[L]$ as {\em justification}.
SGGS employs {\em assignment functions} to keep track of the
{\em dependencies} of $I$-true literals on selected $I$-false literals,
realizing a sort of
{\em first-order propagation modulo semantic guidance} by $I$.
SGGS ensures that $I$-all-true clauses in $\Gamma$
are either conflict clauses or justifications.

The main inference rule of SGGS, called {\em SGGS-extension},
uses the current clause sequence $\Gamma$ and a clause $C$ in $S$
to generate an instance $E$ of $C$ and add it to $\Gamma$ to obtain
the next clause sequence $\Gamma^\prime$.
SGGS-extension is a hyperinference,
because it unifies literals $L_1,\ldots,L_n$ of $C$ with
$I$-false selected literals $M_1,\ldots,M_n$ of opposite sign in $\Gamma$.
The hyperinference is {\em guided} by $I[\Gamma]$,
because $I$-false selected literals contribute to $I[\Gamma]$ as explained above.
Another ingredient of the
instance generation mechanism ensures that every literal in $E$
is either $I$-true or $I$-false.
SGGS-extension is also responsible for selecting a literal in $E$.

The {\em lifting theorem} for SGGS-extension shows that if
$I[\Gamma]\not\models C^\prime$ for some ground instance $C^\prime$
of a clause $C\in S$,
SGGS-extension builds an instance $E$ of $C$ such that $C^\prime$ is an instance of $E$.
There are three kinds of SGGS-extension:
(1) add a clause $E$ which is in conflict with $I[\Gamma]$
and is $I$-all-true;
(2) add a clause $E$ which is in conflict with $I[\Gamma]$
but is not $I$-all-true; and
(3) add a clause $E$ which is not in conflict with $I[\Gamma]$.
In cases (1) and (2), it is necessary to {\em solve the conflict}:
it is here that SGGS lifts the conflict-driven clause learning (CDCL)
mechanism of DPLL-CDCL to the first-order level.

In DPLL-CDCL a conflict is {\em explained}
by resolving a conflict clause $C$
with the justification $D$ of a literal whose complement is in $C$,
generating a new conflict clause.
Typically resolution continues until we get either the empty clause $\perp$
or an {\em asserting clause},
namely a clause where only one literal $Q$ is falsified
in the current decision level.
DPLL-CDCL learns the asserting clause and backjumps to the shallowest level
where $Q$ is undefined
and all other literals in the asserting clause are false,
so that $Q$ enters the trail with the asserting clause as justification.
SGGS {\em explains} a conflict by resolving the conflict clause $E$
with an $I$-all-true clause $D[M]$ in $\Gamma$
which is the justification of the literal $M$
that makes an $I$-false literal $L$ in $E$ uniformly false in $I[\Gamma]$.
Resolution continues until we get either $\perp$
or a conflict clause $E[L]$ which is $I$-all-true.
If $\perp$ arises, $S$ is unsatisfiable.
Otherwise, SGGS {\em moves} the $I$-all-true clause $E[L]$
to the left of the clause $B[M]$,
whose $I$-false selected literal $M$ makes $L$ uniformly false in $I[\Gamma]$.
The effect is to {\em flip} at once the truth value
of {\em all} ground instances of $L$
in $I[\Gamma]$,
so that the conflict is solved, $L$ is implied, and $E[L]$ satisfied.

In order to simplify the presentation,
up to here we omitted that clauses in SGGS may have {\em constraints}.
For example,
$x\not\equiv y \rhd P(x,y) \vee Q(y,x)$ is a {\em constrained clause},
which represents its ground instances that satisfy the constraints:
$P(a,b) \vee Q(b,a)$ is an instance, while $P(a,a) \vee Q(a,a)$ is not.
The reason for constraints is that
selected literals of clauses in $\Gamma$ may {\em intersect},
in the sense of having ground instances with the same atoms.
Since selected literals determine $I^p(\Gamma)$,
whence $I[\Gamma]$,
non-empty intersections represent {\em duplications},
if the literals have the same sign,
and {\em contradictions},
otherwise.
SGGS removes duplications by deletion of clauses,
and contradictions by resolution.
However, before doing either,
it needs to {\em isolate} the shared ground instances
in the selected literal of {\em one} clause.
For this purpose,
SGGS features inference rules that replace a clause by a {\em partition},
that is, a set of clauses that represent the same ground instances
and have {\em disjoint} selected literals.
This requires constraints.
For example,
a partition of $[P(x,y)]\vee Q(x,y)$ is
$\{true \rhd [P(f(z),y)]\vee Q(f(z),y),
\ top(x) \neq f \rhd [P(x,y)]\vee Q(x,y)\}$,
where the constraint $top(x) \neq f$ means that variable $x$
cannot be instantiated with a term whose topmost symbol is $f$.
If $L$ and $M$ in $C[L]$ and $D[M]$ of $\Gamma$ intersect,
SGGS partitions $C[L]$ by $D[M]$:
it partitions $C[L]$ into $A_1 \rhd C_1[L_1],\ldots,A_n \rhd C_n[L_n]$
so that only $L_j$, for some $j$, $1\le j\le n$, intersects with $M$,
and $A_j \rhd C_j[L_j]$ is either deleted or resolved with $D[M]$.

The following example shows an SGGS-refutation:

\begin{example}
Given
$S = \{\neg P(f(x)) \vee \neg Q(g(x)) \vee R(x),\ P(x),\ Q(y),\ \neg R(c)\}$,
let $I$ be all negative.
An SGGS-derivation starts with the empty sequence.
Then, four SGGS-extension steps apply:
$$
\begin{array}{ll}
\Gamma_0\colon & \varepsilon \\
\Gamma_1\colon & [P(x)] \\
\Gamma_2\colon & [P(x)],\ [Q(y)] \\
\Gamma_3\colon & [P(x)],\ [Q(y)],
\ \neg P(f(x)) \vee \neg Q(g(x)) \vee [R(x)] \\
\Gamma_4\colon & [P(x)],\ [Q(y)],
\ \neg P(f(x)) \vee \neg Q(g(x)) \vee [R(x)],\ [\neg R(c)]
\end{array}
$$
At this stage,
the selected literals $R(x)$ and $\neg R(c)$ intersect,
and therefore SGGS partitions $\neg P(f(x)) \vee \neg Q(g(x)) \vee [R(x)]$
by $[\neg R(c)]$:
$$\begin{array}{ll}
\Gamma_5\colon & [P(x)],\ [Q(y)],
\ x \not\equiv c \rhd \neg P(f(x)) \vee \neg Q(g(x)) \vee [R(x)],\\
 & \ \neg P(f(c)) \vee \neg Q(g(c)) \vee [R(c)],\ [\neg R(c)]
\end{array}
$$
Now the $I$-all-true clause $\neg R(c)$ is in conflict with $I[\Gamma_5]$.
Thus, SGGS moves it left of the clause
$\neg P(f(c)) \vee \neg Q(g(c)) \vee [R(c)]$ that makes $\neg R(c)$
false in $I[\Gamma_5]$, in order to amend the induced interpretation.
Then, it resolves these two clauses,
and replaces the parent that is not $I$-all-true,
namely $\neg P(f(c)) \vee \neg Q(g(c)) \vee [R(c)]$,
by the resolvent $\neg P(f(c)) \vee \neg Q(g(c))$:
$$
\begin{array}{ll}
\Gamma_6\colon & [P(x)],\ [Q(y)],
\ x \not\equiv c \rhd \neg P(f(x)) \vee \neg Q(g(x)) \vee [R(x)],
\ [\neg R(c)],\\
 & \neg P(f(c)) \vee \neg Q(g(c)) \vee [R(c)] \\
\Gamma_7\colon & [P(x)],\ [Q(y)],
\ x \not\equiv c \rhd \neg P(f(x)) \vee \neg Q(g(x)) \vee [R(x)],
\ [\neg R(c)],\\
 & \neg P(f(c)) \vee [\neg Q(g(c))]
\end{array}
$$
Assuming that in the resolvent the literal $\neg Q(g(c))$ gets selected,
there is now an intersection between selected literals $\neg Q(g(c))$
and $Q(y)$, so that SGGS partitions $Q(y)$ by
$\neg P(f(c)) \vee \neg Q(g(c))$:
$$
\begin{array}{ll}
\Gamma_8\colon & [P(x)],\ top(y) \neq g \rhd [Q(y)],
\ z \not\equiv c \rhd [Q(g(z))],\ [Q(g(c))],\\
 & x \not\equiv c \rhd \neg P(f(x)) \vee \neg Q(g(x)) \vee [R(x)],
\ [\neg R(c)],\ \neg P(f(c)) \vee [\neg Q(g(c))]
\end{array}
$$
At this point,
the $I$-all-true clause $\neg P(f(c)) \vee [\neg Q(g(c))]$
is in conflict with $I[\Gamma_8]$.
As before, SGGS moves it left of the clause that makes its selected
literal $\neg Q(g(c))$ false, namely $[Q(g(c))]$,
in order to fix the candidate model,
and then resolves $\neg P(f(c)) \vee [\neg Q(g(c))]$
and $[Q(g(c))]$, replacing the latter by the resolvent $\neg P(f(c))$:
$$
\begin{array}{ll}
\Gamma_9\colon & [P(x)],\ top(y) \neq g \rhd [Q(y)],
\ z \not\equiv c \rhd [Q(g(z))],
\ \neg P(f(c)) \vee [\neg Q(g(c))],\\
 & [Q(g(c))],\ x\not\equiv c \rhd \neg P(f(x))\vee\neg Q(g(x))\vee [R(x)],
\ [\neg R(c)]\\
\Gamma_{10}\colon & [P(x)],\ top(y) \neq g \rhd [Q(y)],
\ z \not\equiv c \rhd [Q(g(z))],
\ \neg P(f(c)) \vee [\neg Q(g(c))],\\
 & [\neg P(f(c))],
\ x \not\equiv c \rhd \neg P(f(x)) \vee \neg Q(g(x)) \vee [R(x)],
\ [\neg R(c)]
\end{array}
$$
The resolvent has only one literal which gets selected;
since $[\neg P(f(c))]$ intersects with $[P(x)]$,
the next inference partitions $[P(x)]$ by $[\neg P(f(c))]$:
$$
\begin{array}{ll}
\Gamma_{11}\colon & top(x) \neq f \rhd [P(x)],
\ y \not\equiv c \rhd [P(f(y))],\ [P(f(c))],
\ top(y) \neq g \rhd [Q(y)],\\
 & z \not\equiv c \rhd [Q(g(z))],
\ \neg P(f(c)) \vee [\neg Q(g(c))],\ [\neg P(f(c))],\\
 & x \not\equiv c \rhd \neg P(f(x)) \vee \neg Q(g(x)) \vee [R(x)],
\ [\neg R(c)]
\end{array}
$$
The next step moves the $I$-all-true clause $[\neg P(f(c))]$,
which is in conflict with $I[\Gamma_{11}]$,
to the left of the clause $[P(f(c))]$ that makes $[\neg P(f(c))]$
false in $I[\Gamma_{11}]$,
and then resolves these two clauses
to generate the empty clause:
$$
\begin{array}{ll}
\Gamma_{12}\colon & top(x) \neq f \rhd [P(x)],
\ y \not\equiv c \rhd [P(f(y))],\ [\neg P(f(c))],
\ [P(f(c))],\\
 & top(y) \neq g \rhd [Q(y)],\ z \not\equiv c \rhd [Q(g(z))],
\ \neg P(f(c)) \vee [\neg Q(g(c))],\\
 & x \not\equiv c \rhd \neg P(f(x)) \vee \neg Q(g(x)) \vee [R(x)],
\ [\neg R(c)]\\
\Gamma_{13}\colon & top(x) \neq f \rhd [P(x)],
\ y \not\equiv c \rhd [P(f(y))],\ [\neg P(f(c))],\ \perp,\\
 & top(y) \neq g \rhd [Q(y)],\ z \not\equiv c \rhd [Q(g(z))],
\ \neg P(f(c)) \vee [\neg Q(g(c))],\\
 & x \not\equiv c \rhd \neg P(f(x)) \vee \neg Q(g(x)) \vee [R(x)],
\ [\neg R(c)]
\end{array}
$$
\end{example}

This example only illustrates the basic mechanisms of SGGS.
This method is so new that it has not yet been implemented:
the hope is that its conflict-driven model-repair mechanism
will have on first-order theorem proving an effect similar
to that of the transition from DPLL to DPLL-CDCL for SAT-solvers.
If this were true, even in part,
the benefit could be momentous,
considering that CDCL played a key r\^ole in the success
of SAT technology.
Another expectation is that non-trivial semantic guidance
(i.e., not based on sign like in hyperesolution)
pays off in case of many axioms or large knowledge bases.

\section{Model-based Reasoning in First-Order Theories}\label{mbr-th}

There are basically two ways one can think about a theory
presented by a set of axioms:
as the set of all theorems that are logical consequences of the axioms,
or as the set of all interpretations that are models of the axioms.
The two are obviously connected,
but may lead to different styles of reasoning,
that we portray by the selection of methods in this section.
We cover approaches that {\em build axioms} into resolution,
{\em hierarchical} and {\em locality-based} theory reasoning,
and a recent method called {\em Model-Constructing satisfiability calculus} or {\em MCsat}.

\subsection{Building Theory Axioms into Resolution and Superposition}

The early approaches to theory reasoning emphasized the axioms,
by building them into the inference systems.
The first analyzed theory was {\em equality}:
since submitting the equality axioms to resolution,
or other inference systems for first-order logic,
leads to an explosion of the search space,
{\em paramodulation}, {\em superposition}, and {\em rewriting}
were developed to build equality into resolution
(e.g., \cite{param,HsiRusi:JACM:1991,Rusi:JSC:1991,BacGan:JLC:1994,MPB-JH:TCS:1995:completion}
and Chapters 7 and 9 in \cite{DBLP:books/el/RobinsonV01}).

Once equality was conquered,
research flourished on building-in theories (e.g.,
\cite{plotkin,Peterson-Stickel81,Jouannaud-Kirchner86,HsiRuSa:1987,DerJou:1990,GallierSnyder:1990,JouKir:1991,BdlTE:IC:2007}).
{\em Equational theories},
that are axiomatized by sets of equalities,
and among them {\em permutative theories},
where the two sides of each axiom are permutations of the same symbols,
as in {\em associativity} and {\em commutativity},
received the most attention.
A main ingredient is to replace syntactic unification by
unification {\em modulo} a set $E$ of equational axioms,
a concept generalized by Jos\'e Meseguer to {\em order-sorted $E$-unification}
(e.g., \cite{Meseguer:1989,Meseguer:2009,Meseguer:2012}).
This kind of approach was pursued further,
by building into superposition
axioms for {\em monoids} \cite{waldmann-monoids},
{\em groups} \cite{waldmann-groups},
{\em rings} and {\em modules} \cite{stuber},
or by generalizing superposition to embed {\em transitive relations} other than equality
\cite{BacGan:JACM:1998}.
The complexities and limitations of these techniques led to investigate
the methods for {\em hierarchical} theory reasoning that follow.

\subsection{Hierarchical Reasoning by Superposition}

Since Jos\'e Meseguer's work with Joe Goguen (e.g., \cite{Meseguer:1992}),
it became clear that a major issue at the cross-roads of reasoning,
specifying, and programming,
is that theories, or specifications,
are built by {\em extension} to form {\em hierarchies}.
A {\em base theory} ${\cal T}_0$ is
defined by a set of {\em sorts} ${\cal S}_0$,
a {\em signature} $\Sigma_0$,
possibly a set of axioms $N_0$,
and the class ${\cal C}_0$ of its {\em models}
(e.g., term-generated $\Sigma_0$-algebras).
An {\em extended} or {\em enriched} theory ${\cal T}$ adds new sorts
(${\cal S}_0 \subseteq {\cal S}$),
new function symbols ($\Sigma_0 \subseteq \Sigma$),
called {\em extension functions},
and new axioms ($N_0 \subseteq N$),
specifying properties of the new symbols.
For the base theory the class of models is given,
while the extension is defined axiomatically.
A pair $({\cal T}_0,{\cal T})$ as above forms a {\em hierarchy} with
{\em enrichment axioms} $N$.

The crux of extending specifications was popularized by Joe Goguen and Jos\'e Meseguer
as {\em no junk and no confusion}:
an interpretation of ${\cal S}$ and $\Sigma$,
which is a model of $N$,
is a model of ${\cal T}$ only if
it extends a model in ${\cal C}_0$,
without collapsing its sorts,
or making distinct elements equal ({\em no confusion}),
or introducing new elements of base sort ({\em no junk}).
A sufficient condition for the latter is {\em sufficient completeness},
a property studied also in inductive theorem proving,
which basically says that 
every ground non-base term $t^\prime$ of base sort is equal to a ground base term $t$.
Sufficient completeness is a strong restriction,
violated by merely adding a constant symbol:
if $\Sigma_0 = \{a, b\}$,
$N = N_0 = \{a \not\simeq b\}$,
and $\Sigma = \{a, b, c\}$,
where $a$, $b$, and $c$ are constants of the same sort,
the extension is not sufficiently complete,
because $c$ is junk,
or a model with three distinct elements is not isomorphic to one with two.
Although sufficient completeness is undecidable in general (e.g., \cite{KapurEtAl:1991}),
sufficient completeness analyzers exist
(e.g., \cite{Kapur:tool:1994,Meseguer:2005,Meseguer:2006}),
with key contributions by Jos\'e Meseguer.

{\em Hierarchic superposition} was introduced in \cite{BachmairGanzingerWaldmann}
and developed in \cite{Ganzinger-Waldmann-Sofronie}
to reason about a hierarchy $({\cal T}_0,{\cal T})$ with enrichment 
axioms $N$, where $N$ is a set of clauses.
We assume to have a decision procedure
to detect that a finite set of $\Sigma_0$-clauses is ${\cal T}_0$-unsatisfiable.
Given a set $S$ of $\Sigma$-clauses,
the problem is to determine
whether $S$ is false in all models of the hierarchic specification,
or, equivalently,
whether $N\cup S$ has no model whose reduct to $\Sigma_0$ is a model of ${\cal T}_0$.
The problem is solved by using
the ${\cal T}_0$-reasoner as a black-box to take care of the base part,
while superposition-based inferences apply only to non-base literals.\footnote{Other
approaches to subdivide work between superposition and an SMT-solver appeared in
\cite{MPB-ME:JSC:2010:dpByStages,MPB-CL-LM:JAR:2011:dpllSPsi}.}
First, for every clause $C$, 
whenever a subterm $t$ whose top symbol is a base operator occurs 
immediately below a non-base operator symbol (or vice versa),
$t$ is replaced by a new variable $x$ and the 
equation $x \simeq t$ is added to the antecedent of $C$.
This transformation is called {\em abstraction}.
Then, the inference rules are modified to require that all substitutions are {\em simple},
meaning that they map variables of base sort to base terms. 
A meta-rule named {\em constraint refutation}
detects that a finite set of $\Sigma_0$-clauses is inconsistent in ${\cal T}_0$
by invoking the ${\cal T}_0$-reasoner.
Hierarchic superposition was proved refutationally complete
in \cite{BachmairGanzingerWaldmann},
provided ${\cal T}_0$ is compact,
which is a basic preliminary to make constraint refutation mechanizable,
and $N \cup S$ is {\em sufficiently complete with respect to simple instances},
which means that for every model $I$ 
of all simple ground instances of the clauses in $N \cup S$,
and every ground non-base term $t'$,
there exists a ground base term $t$ (which may depend on $I$)
such that  $I\models t^\prime \simeq t$. 

There are situations where the enrichment adds {\em partial} functions:
$\Sigma_0$ contains only total function symbols,
while $\Sigma\setminus\Sigma_0$ may contain
partial functions and total functions having as codomain a new sort. 
Hierarchic superposition was generalized to handle
both total and partial function symbols,
yielding a {\em partial hierarchic superposition calculus}
\cite{Ganzinger-Waldmann-Sofronie}.
To have an idea of the difficulties posed by partial functions,
consider that replacement of equals by equals may be unsound in their presence.
For example,
$s \not\simeq s$ may hold in a partial algebra
(i.e., a structure where some function symbols are interpreted as partial),
if $s$ is undefined.
Thus, the equality resolution rule
(e.g., resolution between $C \vee s \not\simeq s$ and $x \simeq x$)
is restricted to apply only if $s$ is guaranteed to be defined.
Other restrictions impose that terms replaced 
by inferences may contain a partial function symbol only at the top;
substitutions cannot introduce partial function symbols;
and every ground term made only of total symbols 
is smaller than any ground term containing a partial function symbol
in the ordering used by the inference system.
The following example portrays the partial function case:

\begin{example}
Let ${\cal T}_0$ be the base theory defined by
${\cal S}_0 = \{ {\sf data} \}$,
$\Sigma_0 = \{ b \colon \rightarrow {\sf data},
f \colon {\sf data} \rightarrow {\sf data} \}$,
and $N_0 = \{\forall x \, f(f(x)) \simeq f(x)\}$.
We consider the extension with a new sort ${\sf list}$,
total functions 
$\{{\sf cons} \colon {\sf data}, {\sf list} \rightarrow {\sf list}, 
{\sf nil} \colon \rightarrow {\sf list}, d : \rightarrow {\sf list} \}$,
partial functions
$\{{\sf car} : {\sf list} \rightarrow {\sf data},
{\sf cdr} : {\sf list} \rightarrow {\sf list} \}$,
and the following clauses,
where $N = \{(1), (2), (3)\}$ and $S = \{(4), (5)\}$: 
\[ \begin{array}{lr@{}c@{}l} 
(1) & {\sf car}({\sf cons}(x, l)) & \simeq & x \\
(2) & {\sf cdr}({\sf cons}(x, l)) & \simeq & l  \\
(3) & {\sf cons}({\sf car}(l), {\sf cdr}(l)) & \simeq & l  \\
(4) & f(b) & \simeq & b  \\
(5) & f(f(b)) & \not\simeq & {\sf car}({\sf cdr}({\sf cons}(f(b), {\sf cons}(b, d)))) 
\end{array}\] 
The partial hierarchic superposition calculus deduces: 
{\footnotesize
\begin{tabbing}
(6) \= $x \not\simeq f(f(b)) \vee y \not\simeq f(b) \vee z \not\simeq b
\vee x \not\simeq {\sf car}({\sf cdr}({\sf cons}(y, {\sf cons}(z, d))))$  \= Abstr. (5) \\ 
(7) \> $x \not\simeq f(f(b)) \vee y \not\simeq f(b) \vee z \not\simeq b 
\vee x \not\simeq {\sf car}({\sf cons}(z, d))$  \>  Superp. (2),(6) \\
(8) \> $x \not\simeq f(f(b)) \vee y \not\simeq f(b) \vee z \not\simeq b 
\vee x \not\simeq z$  \>  Superp. (1),(7) \\
(9) \> $\perp$ \> $\!\!\!\!\!\!\!\!\!\!\!\!\!\!\!\!\!\!\!\!\!\!\!\!\!\!\!\!\!\!\!\!\!\!\!\!$Constraint refutation (4),(8)
\end{tabbing}}
\end{example}

Under the assumption that ${\cal T}_0$ is a universal first-order theory,
which ensures compactness,
the partial hierarchic superposition calculus was proved sound and complete
in \cite{Ganzinger-Waldmann-Sofronie}:
if a contradiction cannot be derived from $N\cup S$ using this calculus,
then $N\cup S$ has a model which is a partial algebra. 
Thus, if the unsatisfiability of $N\cup S$ does not depend on the totality
of the extension functions,
the partial hierarchic superposition calculus can detect its inconsistency. 
In certain problem classes where partial algebras can always be made total,
the calculus is complete also for total functions. 
Research on hierarchic superposition continued in \cite{AKW:2009},
where an implementation for extensions of linear arithmetic was presented,
and in \cite{BaumgartnerWaldmann-cade2013},
where the calculus was made ``more complete'' in practice.

\subsection{Hierarchical Reasoning in Local Theory Extensions}

A series of papers starting with \cite{sofronie-cade-05} identified
a class of theory extensions $({\cal T}_0, {\cal T})$,
called {\em local},
which admit a complete hierarchical method for checking satisfiability of {\em ground} clauses,
without requiring either sufficient completeness
or that ${\cal T}_0$ is a universal first-order theory.
The enrichment axioms in $N$ do not have to be clauses:
if they are, 
we have an extension \emph{with clauses};
if $N$ consists of formul\ae\ of the form 
$\forall \bar{x} \, (\Phi(\bar{x}) {\vee} D(\bar{x}))$,
where $\Phi(\bar{x})$ is an \emph{arbitrary} $\Sigma_0$-formula and
$D(\bar{x})$ is a $\Sigma$-clause,
with at least one occurrence of an extension function,
we have an extension \emph{with augmented clauses}.
The basic assumption that ${\cal T}_0$,
or a fragment thereof,
admits a decision procedure for satisfiability clearly remains.

As we saw throughout this survey,
instantiating universally quantified variables is crucial
in first-order reasoning.
Informally,
a theory extension is {\em local},
if it is sufficient to consider only a {\em finite} set of instances.
Let $G$ be a set of ground clauses to be refuted in ${\cal T}$,
and let $N[G]$ denote the set of instances of the clauses in $N$
where every term whose top symbol is an extension function
is a ground term occurring in $N$ or $G$.
Theory ${\cal T}$ is a {\em local extension} of ${\cal T}_0$,
if $N[G]$ suffices to prove the ${\cal T}$-unsatisfiability
of $G$ \cite{sofronie-cade-05}.
Subsequent papers studied variants of locality,
including those for extensions with augmented clauses,
and for combinations of local theories,
and proved that locality can be 
recognized by showing that certain partial algebras embed into total ones
\cite{sofronie-cade-05,sofronie-frocos07,ihlemann-jacobs-sofronie-tacas08,ihlemann-sofronie-ijcar2010}.  

If ${\cal T}$ is a local extension,
it is possible to check the ${\cal T}$-satisfiability of $G$
by hierarchical reasoning
\cite{sofronie-cade-05,sofronie-frocos07,ihlemann-jacobs-sofronie-tacas08,ihlemann-sofronie-ijcar2010},
allowing the introduction of new constants by {\em abstraction} as in
\cite{NelsonOppen:1979}. 
By locality, $G$ is ${\cal T}$-unsatisfiable
if and only if there is no model of $N[G] \cup G$ whose restriction to
$\Sigma_0$ is a model of ${\cal T}_0$.
By abstracting away non-base terms,
$N[G] \cup G$ is transformed into an equisatisfiable set
$N_0 \cup G_0 \cup D$,
where $N_0$ and $G_0$ are sets of $\Sigma_0$-clauses,
and $D$ contains the definitions introduced by abstraction,
namely equalities of the form $f(g_1, \ldots, g_n) {\simeq} c$,
where $f$ is an extension function,
$g_1, \ldots, g_n$ are ground terms,
and $c$ is a new constant. 
The problem is reduced to that of testing the ${\cal T}_0$-satisfiability
of $N_0 \cup G_0 \cup {\sf Con}_0$,
where ${\sf Con}_0$ contains the instances of the congruence axioms
for the terms in $D$:
\[ \displaystyle{{\sf Con}_0  =
\{ \bigwedge_{i = 1}^n c_i \simeq d_i \Rightarrow c \simeq d \mid f(c_1, \dots, c_n) \simeq c, 
 f(d_1, \dots, d_n)\simeq d \in D  \}},\]
which can be solved by a decision procedure
for ${\cal T}_0$ or a fragment thereof.

In the following example ${\cal T}_0$ is the theory of 
{\em linear arithmetic} over the real numbers,
and ${\cal T}$ is its extension with a monotone unary function $f$,
which is known to be a local extension \cite{sofronie-cade-05}:

\begin{example}
Let $G$ be $(a \leq b \wedge f(a) = f(b) + 1)$.
The enrichment
$N = \{ x \leq y \Rightarrow f(x) \leq f(y) \}$
consists of the monotonicity axiom.
In order to check whether $G$ is ${\cal T}$-satisfiable,
we compute $N[G]$, omitting the redundant clauses 
$c \leq c \Rightarrow f(c) \leq f(c)$ for $c \in \{ a, b \}$: 
\[ \begin{array}{l}
N[G] = \{ a \leq b \Rightarrow f(a) \leq f(b),
\ b \leq a \Rightarrow f(b) \leq f(a) \}.
\end{array}\]
The application of abstraction to $N[G] \cup G$ yields $N_0 \cup G_0 \cup D$,
where: 
\[ \begin{array}{l}
N_0 = \{ a \leq b \Rightarrow a_1 \leq b_1,\ b \leq a \Rightarrow b_1 \leq a_1 \},
\quad \quad 
G_0 = \{ a \leq b,\ a_1 \simeq b_1 + 1 \},
\end{array} \]
$D = \{ a_1 \simeq f(a),\ b_1 \simeq f(b) \}$,
and $a_1$ and $b_1$ are new constants. 
Thus, ${\sf Con}_0$ is $\{ a \simeq b \Rightarrow a_1 \simeq b_1 \}$.
A decision procedure for linear arithmetic
applied to $N_0 \cup G_0 \cup {\sf Con}_0$ detects unsatisfiability.
\end{example}

\subsection{Beyond SMT: Satisfiability Modulo Assignment and MCsat} 

Like SGGS generalizes conflict-driven clause learning (CDCL)
to first-order logic and Herbrand interpretations,
the {\em Model-Constructing satisfiability calculus},
or {\em MCsat} for short,
generalizes CDCL to decidable fragments of first-order theories
and their models \cite{dMJ:VMCAI2013,JBdM:FMCAD2013}.

Recall that in DPLL-CDCL the trail that represents the candidate partial model
contains only propositional literals;
the inference mechanism that explains conflicts is propositional resolution;
and learnt clauses are made of input atoms.
These three characteristics are true also of
the DPLL(${\cal T}$) paradigm for SMT-solvers \cite{SMTsurvey:2009},
where an {\em abstraction function} maps finitely many
input first-order ground atoms to finitely many propositional atoms.
In this way,
the method bridges the gap between the first-order language of the theory
${\cal T}$ and the propositional language of the DPLL-CDCL core solver.
In DPLL(${\cal T}$),
also ${\cal T}$-lemmas are made of input atoms,
and the guarantee that no new atoms are generated
is a key ingredient of the proof of termination
of the method in \cite{NOT:2006}.

Also when ${\cal T}$ is a union of theories
${\cal T} = \bigcup_{i=1}^n {\cal T}_i$,
the language of atoms remains finite.
The standard method to combine satisfiability procedures for theories
${\cal T}_1,\ldots,{\cal T}_n$ to get a satisfiability procedure
for their union is {\em equality sharing} \cite{NelsonOppen:1979},
better known as {\em Nelson-Oppen scheme},
even if equality sharing was the original name given by Greg Nelson,
as reconstructed in \cite{Nelson:1983}.
Indeed, a key feature of equality sharing is that the combined procedures
only need to share equalities between constant symbols.
These equalities are mapped by the abstraction function to
{\em proxy variables},
that is, propositional variables that stand for the equalities.
As there are finitely many constant symbols,
there are also finitely many proxy variables.

MCsat generalizes both model representation and inference mechanism
beyond satisfiability modulo theories (SMT),
because it is designed to decide a more general problem
called {\em satisfiability modulo assignment} (SMA).
An SMA problem consists of determining the satisfiability
of a formula $S$ in a theory ${\cal T}$,
given an initial assignment $I$ to some of the variables occuring in $S$,
including {\em both} propositional variables and
{\em free first-order variables}.
SMT can be seen as a special case of SMA where $I$ is empty.
Also, since an SMT-solver builds partial assignments
during the search for a satisfying one,
an intermediate state of an SMT search can be viewed as an instance of SMA.
A first major generalization of MCsat with respect to DPLL-CDCL
and DPLL(${\cal T}$) is to allow the trail to contain also
{\em assignments to free first-order variables} (e.g., $x \gets 3$).
Such assignments can be {\em semantic decisions}
or {\em semantic propagations},
thus called to distinguish them from the Boolean decisions
and Boolean propagations that yield the standard Boolean assignments
(e.g., $L \gets \mathit{true}$).

The answer to an SMA problem is either a model of $S$
including the initial assignment $I$,
or ``unsatisfiable'' with an {\em explanation}, that is,
a formula $S^\prime$ that follows from $S$
and is inconsistent with $I$.
This notion of explanation is a generalization of the explanation of conflicts
by propositional resolution in DPLL-CDCL.
Indeed,
a second major generalization of MCsat with respect to DPLL-CDCL
and DPLL(${\cal T}$)
is to allow the inference mechanism that explains conflicts
to generate {\em new atoms},
as shown in the following example in the
quantifier-free fragment of the theory of equality:

\begin{example}
Assume that $S$ is a conjunction of literals
including $\{v \simeq f(a),\ w \simeq f(b)\}$,
where $a$ and $b$ are constant symbols,
$f$ is a function symbol,
and $v$ and $w$ are free variables.
If the trail contains the assignments
$a\gets\alpha,\ b\gets\alpha,\ w\gets\beta_1,\ v\gets\beta_2$,
where $\alpha$, $\beta_1$, and $\beta_2$ denote distinct values
of the appropriate sorts,
there is a conflict.
The explanation is the formula $a\simeq b \Rightarrow f(a) \simeq f(b)$,
which is an instance of the substitutivity axiom,
or congruence axiom, for function $f$.
Note how the atoms $a\simeq b$ and $f(a) \simeq f(b)$
need not appear in $S$,
and therefore such a lemma could not be generated in DPLL(${\cal T}$).
\end{example}

In order to apply MCsat to a theory ${\cal T}$,
one needs to give clausal inference rules to {\em explain}
conflicts in ${\cal T}$.
These inference rules generate clauses that may contain {\em new}
(i.e., non-input) ground atoms in the signature of the theory.
New atoms come from a {\em basis},
defined as the closure of the set of input atoms
with respect to the inference rules.
The proof of termination of the MCsat transition rules in \cite{dMJ:VMCAI2013}
requires that the basis be {\em finite}.
The following example illustrates the importance of this finiteness
requirement:

\begin{example}\label{ex:MCsat:notGood}
Given $S =
\{x\ge 2,\ \neg (x\ge 1)\vee y\ge 1,\ x^2 + y^2 \le 1 \vee xy > 1\}$,
and starting with an empty trail $M = \emptyset$,
a Boolean propagation puts $x\ge 2$ in the trail.
Theory propagation adds $x\ge 1$,
because $x\ge 2$ implies $x\ge 1$ in the theory,
and $x\ge 1$ appears in $S$.
A Boolean propagation over clause $\neg (x\ge 1)\vee y\ge 1$
adds $y\ge 1$, so that we have $M = x\ge 2,\ x\ge 1,\ y\ge 1$.
If a Boolean decision guesses next $x^2 + y^2 \le 1$
and then a semantic decision adds $x\gets 2$,
we have
$M = x\ge 2,\ x\ge 1,\ y\ge 1,\ x^2 + y^2 \le 1,\ x\gets 2$
and a conflict,
as there is no value for $y$ such that $4 + y^2 \le 1$.
Learning $\neg (x = 2)$ as an explanation of the conflict
does not work, because the procedure can then try $x\gets 3$,
and hit another conflict.
Clearly,
we do not want to learn the infinite sequence
$\neg (x = 2),\ \neg (x = 3),\ \neg (x = 4)\ldots$.
\end{example}

Similarly,
also a systematic application of the inference rules
to enumerate all atoms in a finite basis
would be too inefficient.
The key point is that the inference rules are applied only
to explain conflicts and amend the current partial model,
so that the generation of new atoms is {\em conflict-driven}.
This concept is connected with that of {\em interpolation}
(e.g., \cite{Sofronie:LMCS:2008} for interpolation and locality,
\cite{MPB-MKJ:JAR:2015:interpolationGround} for a survey
on interpolation of ground proofs, and
\cite{MPB-MKJ:JAR:2015:interpolation2Stages}
for an approach to interpolation of non-ground proofs):
given two inconsistent formul\ae\ $A$ and $B$,
a formula that follows from $A$ and is inconsistent with $B$
is an interpolant of $A$ and $B$,
if it is made only of symbols that appear in both $A$ and $B$.
In a theory ${\cal T}$,
the notions of being inconsistent and being logical consequence
are relative to ${\cal T}$,
and the interpolant is allowed to contain theory symbols
even if they are not common to $A$ and $B$.
Since an explanation is a formula $S^\prime$ that follows from $S$
and is inconsistent with $I$,
an interpolant of $S$ and $I$ (written as a formula)
is an explanation.
We illustrate these ideas continuing Example~\ref{ex:MCsat:notGood}:

\begin{example}\label{ex:MCsat:Good}
The solution
is to observe that
$x^2 + y^2 \le 1$ implies $-1 \le x \wedge x \le 1$,
which is inconsistent with $x = 2$.
Note that $-1 \le x \wedge x \le 1$ is an interpolant of
$x^2 + y^2 \le 1$ and $x = 2$,
as $x$ appears in both.
Thus, a desirable explanation is
$(x^2 + y^2 \le 1) \Rightarrow x \le 1$,
or $\neg (x^2 + y^2 \le 1) \vee x \le 1$ in clausal form,
which brings the procedure to update the trail to
$M = x\ge 2,\ x\ge 1,\ y\ge 1,\ x^2 + y^2 \le 1,\ x \le 1$.
At this point, $x\ge 2$ and $x \le 1$ cause another theory conflict,
which leads the procedure to learn the lemma
$\neg (x\ge 2) \vee \neg (x \le 1)$.
A first step of explanation by resolution between 
$\neg (x^2 + y^2 \le 1) \vee x \le 1$ and
$\neg (x\ge 2) \vee \neg (x \le 1)$ yields
$\neg (x^2 + y^2 \le 1) \vee \neg (x\ge 2)$.
A second step of explanation by resolution between
$\neg (x^2 + y^2 \le 1) \vee \neg (x\ge 2)$ and
$x\ge 2$ yields $\neg (x^2 + y^2 \le 1)$,
so that the trail is amended to
$M = x\ge 2,\ x\ge 1,\ y\ge 1,\ \neg (x^2 + y^2 \le 1)$,
finally repairing the decision (asserting $x^2 + y^2 \le 1$)
that caused the conflict.\footnote{The problem in
Examples~\ref{ex:MCsat:notGood} and~\ref{ex:MCsat:Good}
appeared in the slides of a talk entitled
``Arithmetic and Optimization @ MCsat'' presenting joint work
by Leonardo de Moura, Dejan Jovanovi{\'c}, and Grant Olney Passmore,
and given by Leonardo de Moura at a
Schloss Dagsthul Seminar on ``Deduction and Arithmetic'' in October 2013.}
\end{example}

In summary,
MCsat is a fully model-based procedure, 
which lifts CDCL to SMT and SMA.
Assignments to first-order variables and new literals are involved in
decisions, propagations, conflict detections, and explanations,
on a par with Boolean assignments and input literals.
The theories covered in \cite{dMJ:VMCAI2013,JBdM:FMCAD2013} are
the quantifier-free fragments of the theories of
equality, linear arithmetic, and boolean values, and their combinations.
MCsat is also the name of the implementation of the method
as described in \cite{JBdM:FMCAD2013}.

\section{Discussion}

We surveyed model-based reasoning methods,
where inferences build or amend partial models,
which guide in turn further inferences,
balancing search with inference,
and search for a model with search for a proof.
We exemplified these concepts for first-order clausal reasoning,
and then we lifted them, sort of speak, to theory reasoning.
Automated reasoning has made giant strides,
and state of the art systems are very sophisticated
in working with mostly syntactic information.
The challenge of model-based methods is to go
towards a semantically-oriented style of reasoning,
that may pay off for hard problems or new domains.

\subsubsection{Acknowledgments}
The first author thanks David Plaisted,
for starting the research on SGGS and inviting her to join in August 2008;
and Leonardo de Moura,
for the discussions on MCsat at Microsoft Research in Redmond in June 2013.
The third author's work was partially supported by DFG TCRC SFB/TR 14 AVACS
(www.avacs.org).

\bibliography{mbrSurvey-bib}
\bibliographystyle{plain}

\end{document}